\documentclass[10pt,twocolumn,letterpaper]{article}

\usepackage{cvpr}
\usepackage{times}
\usepackage{epsfig}
\usepackage{graphicx}
\usepackage{amsmath}
\usepackage{amssymb}

\usepackage[acronym]{glossaries}
\usepackage{subcaption}
\usepackage{float}
\usepackage{multirow}

\newacronym{SR}{SR}{super-resolution}
\newacronym{LR}{LR}{low-resolution}
\newacronym{HR}{HR}{high-resolution}
\newacronym{CNN}{CNN}{convolutional neural network}
\newacronym{TV}{TV}{total variation}
\newacronym{HD}{HD}{high definition}
\newacronym{MSE}{MSE}{mean squared error}
\newacronym{PSNR}{PSNR}{peak signal-to-noise ratio}
\newacronym{SSIM}{SSIM}{structural similarity}

\usepackage[pagebackref=true,breaklinks=true,letterpaper=true,colorlinks,bookmarks=false]{hyperref}

\usepackage[capitalise]{cleveref}

\cvprfinalcopy 

\DeclareMathOperator*{\argmin}{arg\,min}

\def\cvprPaperID{1989} 

\ifcvprfinal\fi
\begin{document}

\title{Real-Time Video Super-Resolution with Spatio-Temporal Networks and Motion Compensation}


\author{Jose Caballero, Christian Ledig, Andrew Aitken, Alejandro Acosta,\\
Johannes Totz, Zehan Wang, Wenzhe Shi\\
Twitter \\
{\tt\small \{jcaballero, cledig, aaitken, aacostadiaz, johannes, zehanw, wshi\}@twitter.com}
}

\maketitle

\begin{abstract}
Convolutional neural networks have enabled accurate image super-resolution in real-time. However, recent attempts to benefit from temporal correlations in video super-resolution have been limited to naive or inefficient architectures. In this paper, we introduce spatio-temporal sub-pixel convolution networks that effectively exploit temporal redundancies and improve reconstruction accuracy while maintaining real-time speed. Specifically, we discuss the use of early fusion, slow fusion and 3D convolutions for the joint processing of multiple consecutive video frames. We also propose a novel joint motion compensation and video super-resolution algorithm that is orders of magnitude more efficient than competing methods, relying on a fast multi-resolution spatial transformer module that is end-to-end trainable. These contributions provide both higher accuracy and temporally more consistent videos, which we confirm qualitatively and quantitatively. Relative to single-frame models, spatio-temporal networks can either reduce the computational cost by 30\% whilst maintaining the same quality or provide a 0.2dB gain for a similar computational cost. Results on publicly available datasets demonstrate that the proposed algorithms surpass current state-of-the-art performance in both accuracy and efficiency.

\end{abstract}

\section{Introduction}

Image and video \gls{SR} are long-standing challenges of signal processing. \Gls{SR} aims at recovering a \gls{HR} image or video from its \gls{LR} version, and finds direct applications ranging from medical imaging \cite{Yang2012, Shi2013} to satellite imaging \cite{Karunakar2013}, as well as facilitating tasks such as face recognition \cite{Gunturk2003}. The reconstruction of \gls{HR} data from a \gls{LR} input is however a highly ill-posed problem that requires additional constraints to be solved. While those constraints are often application-dependent, they usually rely on data redundancy.

In single image \gls{SR}, where only one \gls{LR} image is provided, methods exploit inherent image redundancy in the form of local correlations to recover lost high-frequency details by imposing sparsity constraints \cite{Yang2010} or assuming other types of image statistics such as multi-scale patch recurrence \cite{Glasner2009}. In multi-image \gls{SR} \cite{Park2003} it is assumed that different observations of the same scene are available, hence the shared explicit redundancy can be used to constrain the problem and attempt to invert the downscaling process directly. Transitioning from images to videos implies an additional data dimension (time) with a high degree of correlation that can also be exploited to improve performance in terms of accuracy as well as efficiency.



\subsection{Related work}

\begin{figure}[t]
\centering
	\includegraphics[width=\columnwidth]{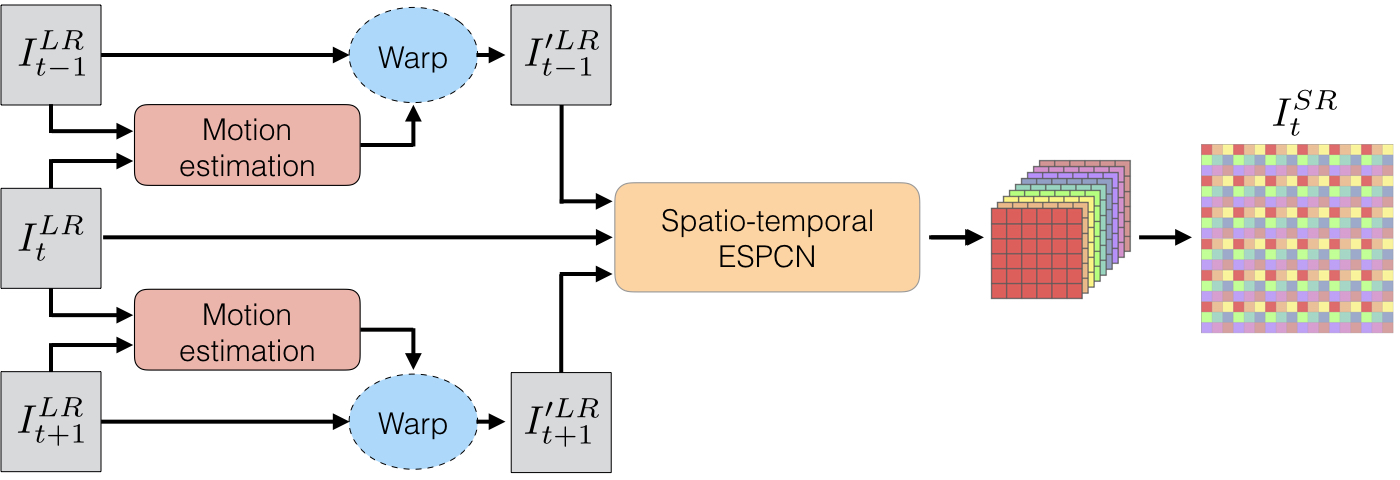}
	\caption{Proposed design for video \gls{SR}. The motion estimation and ESPCN modules are learnt end-to-end to obtain a motion compensated and fast algorithm.}
	\label{fig:network}
\end{figure}

Video \gls{SR} methods have mainly emerged as adaptations of image \gls{SR} techniques. Kernel regression methods \cite{Takeda2007} have been shown to be applicable to videos using 3D kernels instead of 2D ones \cite{Takeda2009}. Dictionary learning approaches, which define \gls{LR} images as a sparse linear combination of dictionary atoms coupled to a \gls{HR} dictionary, have also been adapted from images \cite{Yang2012} to videos \cite{Dai2015}. Another approach is example-based patch recurrence, which assumes patches in a single image or video obey multi-scale relationships, and therefore missing high-frequency content at a given scale can be inferred from coarser scale patches. This was successfully presented by Glasner et al. \cite{Glasner2009} for image \gls{SR} and has later been extended to videos \cite{Shahar2011}.

When adapting a method from images to videos it is usually beneficial to incorporate the prior knowledge that frames of the same scene of a video can be approximated by a single image and a motion pattern. Estimating and compensating motion is a powerful mechanism to further constrain the problem and expose temporal correlations. It is therefore very common to find video \gls{SR} methods that explicitly model motion through frames. A natural choice has been to preprocess input frames by compensating inter-frame motion using displacement fields obtained from off-the-shelf optical flow algorithms \cite{Takeda2009}. This nevertheless requires frame preprocessing and is usually expensive. Alternatively, motion compensation can also be performed jointly with the \gls{SR} task, as done in the Bayesian approach of Liu et al. \cite{Liu2015} by iteratively estimating motion as part of its wider modeling of the downscaling process.

The advent of neural network techniques that can be trained from data to approximate complex nonlinear functions has set new performance standards in many applications including \gls{SR}. Dong et al. \cite{Dong2015} proposed to use a \gls{CNN} architecture for single image \gls{SR} that was later extended by Kappeler et al. \cite{Kappeler2016} in a video \gls{SR} network (VSRnet) which jointly processes multiple input frames. Additionally, compensating the motion of input images with a \gls{TV}-based optical flow algorithm showed an improved accuracy. Joint motion compensation for \gls{SR} with neural networks has also been studied through recurrent bidirectional networks \cite{Huang2015}.

The common paradigm for \gls{CNN} based approaches has been to upscale the \gls{LR} image with bicubic interpolation before attempting to solve the \gls{SR} problem \cite{Dong2015, Kappeler2016}. However, increasing input image size through interpolation considerably impacts the computational burden for \gls{CNN} processing. A solution was proposed by Shi et al. with an efficient sub-pixel convolution network (ESPCN) \cite{Shi2016}, where an upscaling operation directly mapping from \gls{LR} to \gls{HR} space is learnt by the network. This technique reduces runtime by an order of magnitude and enables real-time video \gls{SR} by independently processing frames with a single frame model. Similar solutions  to improve efficiency have also been proposed based on transposed convolutions \cite{Dong2016, Johnson2016}.


\subsection{Motivation and contributions}

\begin{figure*}
  \centering
  \begin{subfigure}[b]{0.32\textwidth}
      \centering
      \fbox{\includegraphics[width=\textwidth]{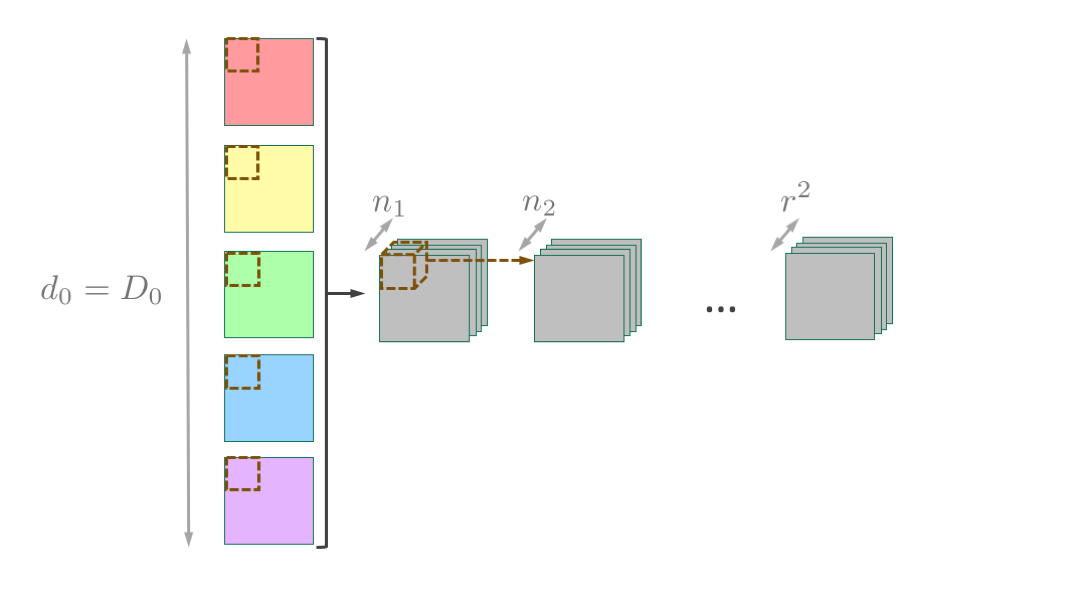}}
      \caption{Early fusion}
      \label{fig:early-fusion}
  \end{subfigure}
  \hspace{\fill}
  \begin{subfigure}[b]{0.32\textwidth}
      \centering
      \fbox{\includegraphics[width=\textwidth]{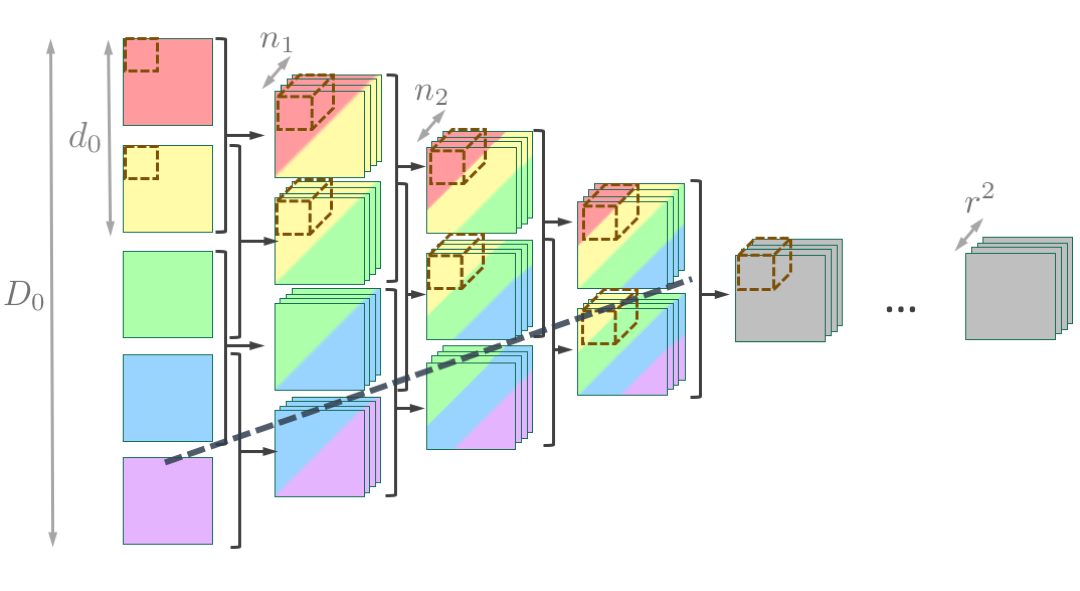}}
      \caption{Slow fusion}
      \label{fig:slow-fusion}
  \end{subfigure}
  \hspace{\fill}
  \begin{subfigure}[b]{0.32\textwidth}
      \centering
      \fbox{\includegraphics[width=\textwidth]{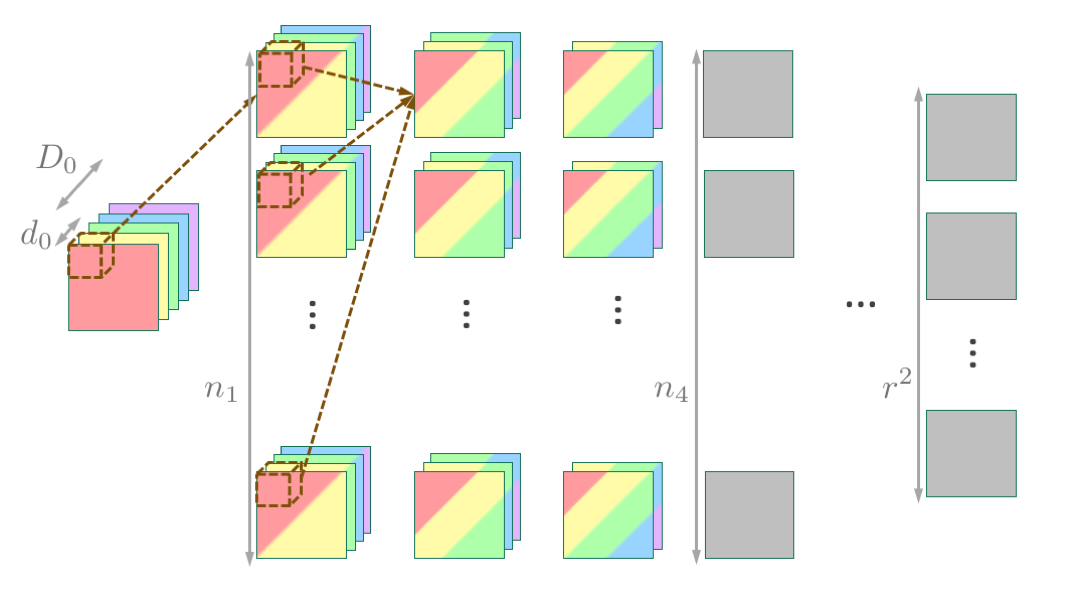}}
      \caption{3D convolution}
      \label{fig:3dconv}
  \end{subfigure}
  \caption{Spatio-temporal models. Input frames are colour coded to illustrate their contribution to different feature maps, and brackets represent convolution after concatenation. In early fusion (a), the temporal depth of the network's input filters matches the number of input frames collapsing all temporal information in the first layer. In slow fusion (b), the first layers merge frames in groups smaller than the input number of frames. If weights in each layer are forced to share their values, operations needed for features above the dashed line can be reused for each new frame. This case is equivalent to using 3D convolutions (c), where the temporal information is merged with convolutions in space and time.}
\label{fig:st-networks}
\end{figure*}

Existing solutions for \gls{HD} video \gls{SR} have not been able to effectively exploit temporal correlations while performing in real-time. On the one hand, ESPCN \cite{Shi2016} leverages sub-pixel convolution for a very efficient operation, but its naive extension to videos treating frames independently fails to exploit inter-frame redundancies and does not enforce a temporally consistent result. VSRnet \cite{Kappeler2016}, on the other hand, can improve reconstruction quality by jointly processing multiple input frames. However, the preprocessing of \gls{LR} images with bicubic upscaling and the use of an inefficient motion compensation mechanism slows runtime to about $0.016$ frames per second even on videos smaller than standard definition resolution.

Spatial transformer networks \cite{Jaderberg2015} provide a means to infer parameters for a spatial mapping between two images. These are differentiable networks that can be seamlessly combined and jointly trained with networks targeting other objectives to enhance their performance. For instance, spatial transformer networks were initially shown to facilitate image classification by transforming images onto the same frame of reference \cite{Jaderberg2015}. Recently, it has been shown how spatial transformers can encode optical flow features with unsupervised training \cite{Ganin2016, Ahmadi2016, Patraucean2016a, Handa2016}, but they have nevertheless not yet been investigated for video motion compensation. Related approaches have emerged for view synthesis assuming rigid transformations \cite{Kalantari2016}.


In this paper, we combine the efficiency of sub-pixel convolution with the performance of spatio-temporal networks and motion compensation to obtain a fast and accurate video \gls{SR} algorithm. We study different treatments of the temporal dimension with early fusion, slow fusion and 3D convolutions, which have been previously suggested to extend classification from images to videos \cite{Karpathy2014a, Tran2015}. Additionally, we build a motion compensation scheme based on spatial transformers, which is combined with spatio-temporal models to lead to a very efficient solution for video \gls{SR} with motion compensation that is end-to-end trainable. A high-level diagram of the proposed approach is show in \cref{fig:network}.

The main contributions of this paper are:

\begin{itemize}
\item Presenting a real-time approach for video \gls{SR} based on sub-pixel convolution and spatio-temporal networks that improves accuracy and temporal consistency.

\item Comparing early fusion, slow fusion and 3D convolutions as alternative architectures for discovering spatio-temporal correlations.

\item Proposing an efficient method for dense inter-frame motion compensation based on a multi-scale spatial transformer network.

\item Combining the proposed motion compensation technique with spatio-temporal models to provide an efficient, end-to-end trainable motion compensated video \gls{SR} algorithm.
\end{itemize}

\section{Methods}

Our starting point is the real-time image \gls{SR} method ESPCN \cite{Shi2016}. We restrict our analysis to standard architectural choices and do not further investigate potentially beneficial extensions such as recurrence \cite{Kim2015a}, residual connections \cite{He2015, He2016} or training networks based on perceptual loss functions \cite{Johnson2016, Ledig2016, Bruna2016, Dosovitskiy2016}. Throughout the paper we assume all image processing is performed on the y-channel in colour space, and thus we represent all images as 2D matrices.

\subsection{Sub-pixel convolution SR}

For a given \gls{LR} image $I^{LR} \in \mathbb{R}^{H \times W}$ which is assumed to be the result of low-pass filtering and downscaling by a factor $r$ the \gls{HR} image $I^{HR} \in \mathbb{R}^{rH \times rW}$, the \gls{CNN} super-resolved solution $I^{SR} \in \mathbb{R}^{rH \times rW}$ can be expressed as 
\begin{equation}\label{eq:image-sr}
I^{SR} = f \left( I^{LR}; \theta \right).
\end{equation}
Here, $\theta$ are model parameters and $f(.)$ represents the mapping function from \gls{LR} to \gls{HR}. A convolutional network models this function as a concatenation of $L$ layers defined by sets of weights and biases $\theta_l = \left( W_l, b_l \right)$, each followed by non-linearities $\phi_l$, with $l \in [0, L-1]$. Formally, the output of each layer is written as
\begin{equation}
f_l \left( I^{LR}; \theta_l \right) = \phi_l \left( W_l \ast f_{l-1} \left( I^{LR}; \theta_{l-1} \right) + b_l \right), \forall l,
\end{equation}
with $f_0 \left( I^{LR}; \theta_0 \right) = I^{LR}$. We assume the shape of filtering weights to be $n_{l-1} \times n_l \times k_l \times k_l$, where $n_l$ and $k_l$ represent the number and size of filters in layer $l$, with the single frame input meaning $n_0 = 1$. Model parameters are optimised minimising a loss given a set of \gls{LR} and \gls{HR} example image pairs, commonly \gls{MSE}:
\begin{equation}\label{eq:image-sr-objective}
\theta^{\ast} = \argmin_{\theta} \| I^{HR} - f(I^{LR}; \theta) \|_2^2.
\end{equation}

Methods preprocessing $I^{LR}$ with bicubic upsampling before mapping from \gls{LR} to \gls{HR} impose that the output number of filters is $n_{L-1} = 1$ \cite{Dong2015, Kappeler2016}. Using sub-pixel convolution allows to process $I^{LR}$ directly in the \gls{LR} space and then use $n_{L-1} = r^2$ output filters to obtain an \gls{HR} output tensor with shape $1 \times r^2 \times H \times W$ that can be reordered to obtain $I^{SR}$ \cite{Shi2016}. This implies that if there exists an upscaling operation that is better suited for the problem than bicubic upsampling, the network can learn it. Moreover, and most importantly, all convolutional processing is performed in \gls{LR} space, making this approach very efficient.

\subsection{Spatio-temporal networks}\label{ssec:st-networks}

Spatio-temporal networks assume input data to be a block of spatio-temporal information, such that instead of a single input frame $I^{LR}$, a sequence of consecutive frames is considered. This can be represented in the network by introducing an additional dimension for temporal depth $D_l$, with the input depth $D_0$ representing an odd number of consecutive input frames. If we denote the temporal radius of a spatio-temporal block to be $R = \frac{D_0-1}{2}$, we define the group of input frames centered at time $t$ as $I^{LR}_{[t-R:t+R]} \in \mathbb{R}^{H \times W \times D_0}$, and the problem in \cref{eq:image-sr} becomes
\begin{equation}\label{eq:video-sr}
I^{SR}_t = f \left( I^{LR}_{[t-R:t+R]}; \theta \right).
\end{equation}
The shape of weighting filters $W_l$ is also extended by their temporal size $d_l$, and their tensor shape becomes $d_l \times n_{l-1} \times n_l \times k_l \times k_l$. We note that it is possible to consider solutions that aim to jointly reconstruct more than a single output frame, which could have advantages at least in terms of computational efficiency. However, in this work we focus on the reconstruction of only a single output frame.


\subsubsection{Early fusion}

One of the most straightforward approaches for a \gls{CNN} to process videos is to match the temporal depth of the input layer to the number of frames $d_0=D_0$. This will collapse all temporal information in the first layer and the remaining operations are identical to those in a single image \gls{SR} network, meaning $d_l=1, l \ge 1$. An illustration of early fusion is shown in \cref{fig:early-fusion} for $D_0=5$, where the temporal dimension has been colour coded and the output mapping to 2D space is omitted. This design has been studied for video classification and action recognition \cite{Karpathy2014a, Tran2015}, and was also one of the architectures proposed in VSRnet \cite{Kappeler2016}. However, VSRnet requires bicubic upsampling as opposed to sub-pixel convolution, making the framework computationally much less efficient in comparison.


\subsubsection{Slow fusion}

Another option is to partially merge temporal information in a hierarchical structure, so it is slowly fused as information progresses through the network. In this case, the temporal depth of network layers is configured to be $1 \le d_l < D_0$, and therefore some layers also have a temporal extent until all information has been merged and the depth of the network reduces to $1$. This architecture, termed slow fusion, has shown better performance than early fusion for video classification \cite{Karpathy2014a}. In \cref{fig:slow-fusion} we show a slow fusion network where $D_0=5$ and the rate of fusion is defined by $d_l=2$ for $l \le 3$ or $d_l=1$ otherwise, meaning that at each layer only two consecutive frames or filter activations are merged until the network's temporal depth shrinks to $1$. Note that early fusion is an special case of slow fusion.


\subsubsection{3D convolutions}

Another variation of slow fusion is to force layer weights to be shared across the temporal dimension, which has computational advantages. Assuming an online processing of frames, when a new frame becomes available the result of some layers for the previous frame can be reused. For instance, refering to the diagram in \cref{fig:slow-fusion} and assuming the bottom frame to be the latest frame received, all activations above the dashed line are readily available because they were required for processing the previous frame. This architecture is equivalent to using 3D convolutions, initially proposed as an effective tool to learn spatio-temporal features that can help for video action recognition \cite{Tran2015}. An illustration of this design from a 3D convolution perspective is shown in \cref{fig:3dconv}, where the arrangement of the temporal and filter features is swapped relative to \cref{fig:slow-fusion}.


\subsection{Spatial transformer motion compensation}\label{sec:transformer}

We propose the use of an efficient spatial transformer network to compensate the motion between frames fed to the \gls{SR} network. It has been shown how spatial transformers can effectively encode optical flow to describe motion \cite{Patraucean2016a, Ahmadi2016, Handa2016}, and are therefore suitable for motion compensation. We will compensate blocks of three consecutive frames to combine the compensation module with the \gls{SR} network as shown in \cref{fig:network}, but for simplicity we first introduce motion compensation between two frames. Notice that the data used contains inherent motion blur and (dis)occlusions, and even though an explicit modelling for these effects is not used it could potentially improve results.

The task is to find the best optical flow representation relating a new frame $I_{t+1}$ with a reference current frame $I_{t}$. The flow is assumed pixel-wise dense, allowing to displace each pixel to a new position, and the resulting pixel arrangement requires interpolation back onto a regular grid. We use bilinear interpolation $\mathcal{I} \{.\}$ as it is much more efficient than the thin-plate spline interpolation originally proposed in \cite{Jaderberg2015}. Optical flow is a function of parameters $\theta_{\Delta, t+1} $ and is represented with two feature maps $\Delta_{t+1} = \left( \Delta_{t+1} x, \Delta_{t+1} y ; \theta_{\Delta,t+1} \right)$ corresponding to displacements for the $x$ and $y$ dimensions, thus a compensated image can be expressed as $I_{t+1}' \left( x, y \right) = \mathcal{I} \{ I_{t+1} \left( x + \Delta_{t+1} x, y + \Delta_{t+1} y \right) \}$, or more concisely
\begin{equation}\label{eq:motion-compensation}
I_{t+1}' = \mathcal{I} \{ I_{t+1}(\Delta_{t+1}) \}.
\end{equation}

\begin{figure}[t]
  \centering
    \includegraphics[width=0.8\columnwidth]{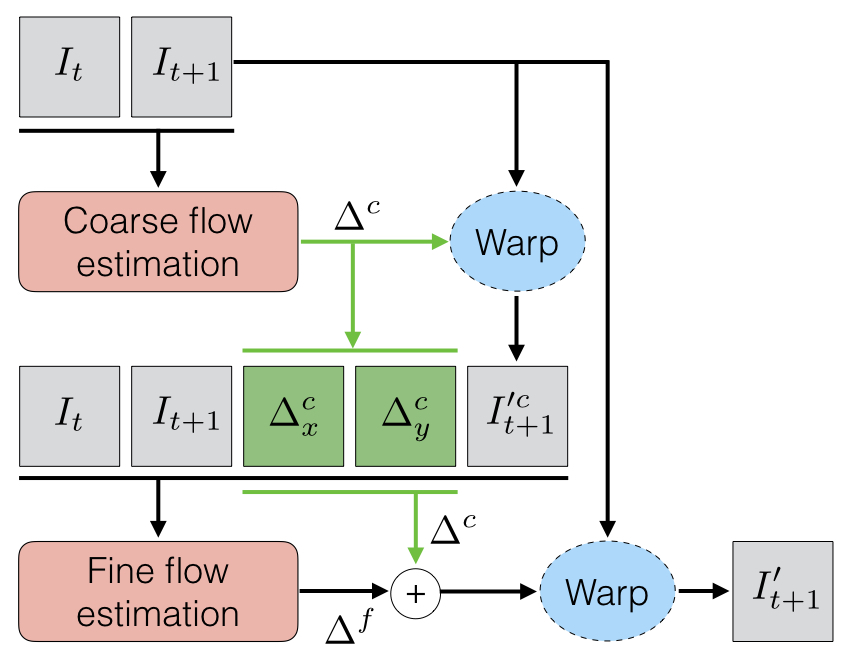}
  \caption{Spatial transformer motion compensation.}
  \label{fig:transformer}
\end{figure}

We adopt a multi-scale design to represent the flow, which has been shown to be effective in classical methods \cite{Farneback2003, Brox2004} and also in more recently proposed spatial transformer techniques \cite{Ganin2016, Ahmadi2016, Fischer2015}. A schematic of the design is shown in \cref{fig:transformer} and flow estimation modules are detailed in \cref{tab:transformer}. First, a $\times 4$ coarse estimate of the flow is obtained by early fusing the two input frames and downscaling spatial dimensions with $\times 2$ strided convolutions. The estimated flow is upscaled with sub-pixel convolution and the result $\Delta^c_{t+1}$ is applied to warp the target frame producing $I'^c_{t+1}$. The warped image is then processed together with the coarse flow and the original images through a fine flow estimation module. This uses a single strided convolution with stride $2$ and a final $\times 2$ upscaling stage to obtain a finer flow map $\Delta^f$. The final motion compensated frame is obtained by warping the target frame with the total flow $I_{t+1}' = \mathcal{I} \{ I_{t+1}(\Delta^c_{t+1} + \Delta^f_{t+1}) \}$. Output activations use tanh to represent pixel displacement in normalised space, such that a displacement of $\pm1$ means maximum displacement from the center to the border of the image.

To train the spatial transformer to perform motion compensation we optimise its parameters $\theta_{\Delta,t+1}$ to minimise the \gls{MSE} between the transformed frame and the reference frame. Similary to classical optical flow methods, we found that it is generally helpful to constrain the flow to behave smoothly in space, and so we penalise the Huber loss of the flow map gradients, namely
\begin{equation}\label{eq:motion-compensation-objective}
\theta_{\Delta, t+1}^*  = \argmin_{\theta_{\Delta,t+1}} \| I_{t} - I'_{t+1} \|_2^2 + \lambda \mathcal{H} \left( \partial_{x,y} \Delta_{t+1} \right).
\end{equation}
In practice we approximate the Huber loss with $\mathcal{H} \left( \partial_{x,y} \Delta \right) = \sqrt{\epsilon + \sum_{i=x,y} (\partial_x \Delta i^2 + \partial_y \Delta i^2 )}$, where $\epsilon = 0.01$. This function has a smooth $L2$ behaviour near the origin and is sparsity promoting far from it. 

The spatial transformer module is advantageous relative to other motion compensation mechanisms as it is straightforward to combine with a \gls{SR} network to perform joint motion compensation and video \gls{SR}. Referring to \cref{fig:network}, the same parameters $\theta_{\Delta}$ can be used to model motion of the outer two frames relative to the central frame. The spatial transformer and \gls{SR} modules are both differentiable and therefore end-to-end trainable. As a result, they can be jointly optimised to minimise a composite loss combining the accuracy of the reconstruction in \cref{eq:image-sr-objective} with the fidelity of motion compensation in \cref{eq:motion-compensation-objective}, namely
\begin{equation}\label{eq:video-sr-memc}
  \begin{split}
    (\theta^*, \theta_{\Delta}^*) & = \argmin_{\theta, \theta_\Delta} \| I^{HR}_t - f(I'^{LR}_{t-1:t+1}; \theta) \|_2^2 \\
&\phantom{=}\, + \sum_{i=\pm 1} [ \beta \| I_{t+i}'^{LR} - I_t^{LR} \|_2^2 + \lambda \mathcal{H} \left( \partial_{x,y} \Delta_{t+i} \right) ].
  \end{split}
\end{equation}
\begin{table}[tb]
\footnotesize
\begin{tabular}{c|l|l}
Layer & \multicolumn{1}{c|}{Coarse flow} & \multicolumn{1}{c}{Fine flow} \\ \hline
1   & Conv k5-n24-s2 / ReLU & Conv k5-n24-s2 / ReLU \\
2   & Conv k3-n24-s1 / ReLU & Conv k3-n24-s1 / ReLU \\
3   & Conv k5-n24-s2 / ReLU & Conv k3-n24-s1 / ReLU \\
4   & Conv k3-n24-s1 / ReLU & Conv k3-n24-s1 / ReLU \\
5   & Conv k3-n32-s1 / tanh & Conv k3-n8-s1 / tanh \\
6   & Sub-pixel upscale $\times 4$ & Sub-pixel upscale $\times 2$
\end{tabular}
\caption{Motion compensation transformer architecture. Convolutional layers are described by kernel size (k), number of features (n) and stride (s).}
\label{tab:transformer}
\end{table}

\section{Experiments and results}

In this section, we first analyse spatio-temporal networks for video \gls{SR} in isolation and later evaluate the benefits of introducing motion compensation. We restrict our experiments to tackle $\times 3$ and $\times 4$ upscaling of full \gls{HD} video resolution ($1080 \times 1920$), and no compression is applied. To ensure a fair comparison of methods, the number of network parameters need to be comparable so that gains in performance can be attributed to specific choices of network resource allocation and not to a trivial increase in capacity. For a layer $l$, the number of floating-point operations to reconstruct a frame is approximated by 
\begin{equation}\label{eq:operations}
HWD_{l+1}n_{l+1} \bigg[ \overbrace{(2 k_l^2 d_l - 1)n_l}^\text{convolutions} + \underbrace{2}_\text{bias \& activation}\bigg].
\end{equation}
In measuring the complexity of slow fusion networks with weight sharing we look at steady-state operation where the output of some layers is reused from one frame to the following. We note that the analysis of VSRnet variants in \cite{Kappeler2016} does not take into account model complexity.

\subsection{Experimental setup}


\subsubsection{Data}

We use the CDVL database \cite{cdvl}, which contains $115$ uncompressed full \gls{HD} videos excluding repeated videos, and choose a subset of $100$ videos for training. The videos are downscaled and $30$ random samples are extracted from each \gls{HR}-\gls{LR} video pair to obtain $3000$ training samples, $5\%$ of which are used for validation. Depending on the network architecture, we refer to a sample as a single input-output frame pair for single frame networks, or as a block of consecutive \gls{LR} input frames and the corresponding central \gls{HR} frame for spatio-temporal networks. The remaining $15$ videos are used for testing. Although the total number of training frames is large, we foresee that the methods presented could benefit from a richer, more diverse set of videos. Additionally, we present a benchmark against various \gls{SR} methods on publicly available videos that are recurrently used in the literature and we refer to as Vid4\footnote{Vid4 is composed of \textit{walk}, \textit{city}, \textit{calendar} and \textit{foliage}, and has sizes $720 \times 480$ or $720 \times 576$. The sequence \textit{city} has dimensions $704 \times 576$, which we crop to $702 \times 576$ for $\times 3$ upscaling. Results on Vid4 can be downloaded from \url{https://twitter.box.com/v/vespcn-vid4}}.

\subsubsection{Network training and parameters}

All \gls{SR} models are trained following the same protocol and share similar hyperparameters. Filter sizes are set to $k_l=3$ $\forall l$, and all non-linearities $\phi_l$ are rectified linear units except for the output layer, which uses a linear activation. Biases are initialised to $0$ and weights use orthogonal initialisation with gain $\sqrt{2}$ following recommendations in \cite{Saxe2013}. All hidden layers are set to have the same number of features. Video samples are broken into non-overlapping sub-samples of spatial dimensions $33 \times 33$, which are randomly grouped in batches for stochastic optimisation. We employ Adam \cite{Kingma2014} with a learning rate $10^{-4}$ and an initial batch size $1$. Every $10$ epochs the batch size is doubled until it reaches a maximum size of $128$.

We choose $n_l=24$ for layers where the network temporal depth is $1$ (layers in gray in \cref{fig:early-fusion,fig:slow-fusion,fig:3dconv}), and to maintain comparable network sizes we choose $n_l=24/D_l, l>0$. This ensures that the number of features per hidden layer in early and slow fusion networks is always the same. For instance, the network shown in \cref{fig:slow-fusion}, for which $D_0=5$ and $d_l = 2$ for $l \le 3$, the number of features in a $6$ layer network for $\times r$ \gls{SR} would be 6, 8, 12, 24, 24, $r^2$.



\subsection{Spatio-temporal video SR}

\subsubsection{Single vs multi frame early fusion}

\begin{figure}[t]
  \centering
    \includegraphics[width=\columnwidth]{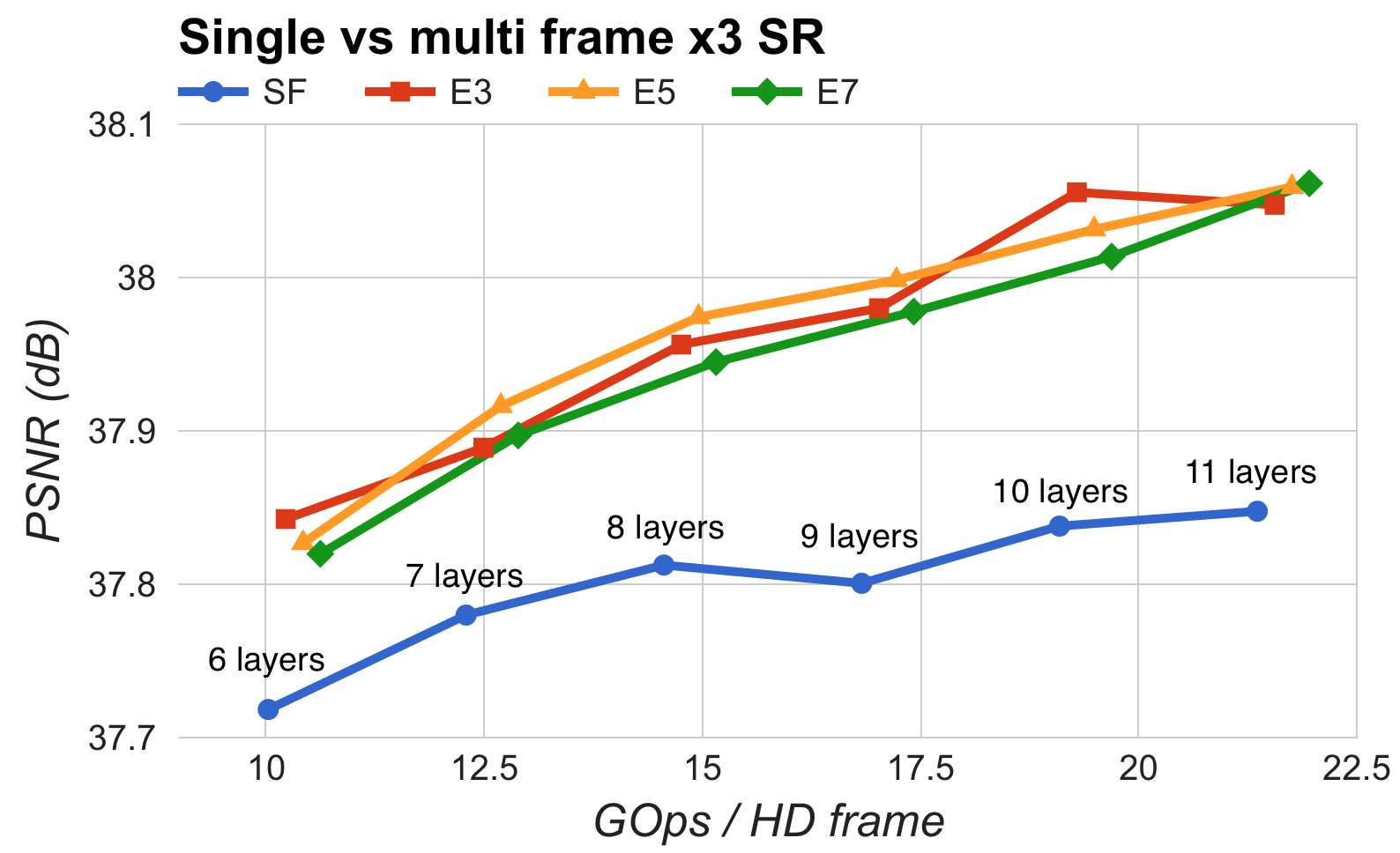}
  \caption{CDVL $\times 3$ \gls{SR} using single frame models (SF) and multi frame early fusion models (E3-7).}
  \label{fig:single-vs-multi-frame}
\end{figure}

First, we investigate the impact of the number of input frames on complexity and accuracy without motion compensation. We compare single frame models (SF) against early fusion spatio-temporal models using 3, 5 and 7 input frames (E3, E5 and E7). \Gls{PSNR} results on the CDVL dataset for networks of 6 to 11 layers are plotted in \cref{fig:single-vs-multi-frame}. Exploiting spatio-temporal correlations provides a more accurate result relative to an independent processing of frames. The increase in complexity from early fusion is marginal because only the first layer contributes to an increase of operations.

Although the accuracy of spatio-temporal models is relatively similar, we find that E7 slightly underperforms. It is likely that temporal dependencies beyond 5 frames become too complex for networks to learn useful information and act as noise degrading their performance. Notice also that, whereas the performance increase from network depth is minimal after 8 layers for single frame networks, this increase is more consistent for spatio-temporal models.

\subsubsection{Early vs slow fusion}

\begin{table}[tb]
\centering
\footnotesize
\begin{tabular}{c|c|cccc}
\# Layers & & SF & E5      & S5      & S5-SW       \\ \hline
\multirow{2}{*}{7}   & PSNR & 37.78 & 37.92 & 37.83 & 37.74  \\
								& GOps & 12.29 & 12.69 & 10.65 & 8.94  \\ \hline
\multirow{2}{*}{9}   & PSNR & 37.80 & 37.99 & 37.99 &  37.90   \\
     							& GOps & 16.83 & 17.22 & 15.19 & 13.47
\end{tabular}
\caption{Comparison of spatio-temporal architectures}
\label{tab:early-vs-slow-fusion}
\end{table}

Here we compare the different treatments of the temporal dimension discussed in \cref{ssec:st-networks}. We assume networks with an input of $5$ frames and slow fusion models with filter temporal depths $2$ as in \cref{fig:st-networks}. Using SF, E5, S5, and S5-SW to refer to single frame networks and 5 frame input networks using early fusion, slow fusion, and slow fusion with shared weights, we show in \cref{tab:early-vs-slow-fusion} results for 7 and 9 layer networks. 

As seen previously, early fusion networks attain a higher accuracy at a marginal 3\% increase in operations relative to the single frame models, and as expected, slow fusion architectures provide efficiency advantages. Slow fusion is faster than early fusion because it uses fewer features in the initial layers. Referring to \cref{eq:operations}, slow fusion uses $d_l=2$ in the first layers and $n_l=24/D_l$, which results in fewer operations than $d_l=1$, $n_l=24$ as used in early fusion.

While the 7 layer network sees a considerable decrease in accuracy using slow fusion relative to early fusion, the 9 layer network can benefit from the same accuracy while reducing its complexity with slow fusion by about 30\%. This suggests that in shallow networks the best use of network resources is to utilise the full network capacity to jointly process all temporal information as done by early fusion, but that in deeper networks slowly fusing the temporal dimension is beneficial, which is in line with the results presented by \cite{Karpathy2014a} for video classification.

Additionally, weight sharing decreases accuracy because of the reduction in network parameters, but the reusability of network features means fewer operations are needed per frame. For instance, the 7 layer S5-SW network shows a reduction of almost 30\% of operations with a minimal decrease in accuracy relative to SF. Using 7 layers with E5 nevertheless shows better performance and faster operation than S5-SW with 9 layers, and in all cases we found that early or slow fusion consistently outperformed slow fusion with shared weights in this performance and efficiency trade-off. Convolutions in spatio-temporal domain were shown in \cite{Tran2015} to work well for video action recognition, but with larger capacity and many more frames processed jointly. We speculate this could be the reason why the conclusions drawn from this high-level vision task do not extrapolate to the \gls{SR} problem.


\subsection{Motion compensated video SR}

\begin{figure}
	\centering
    \includegraphics[width=\columnwidth]{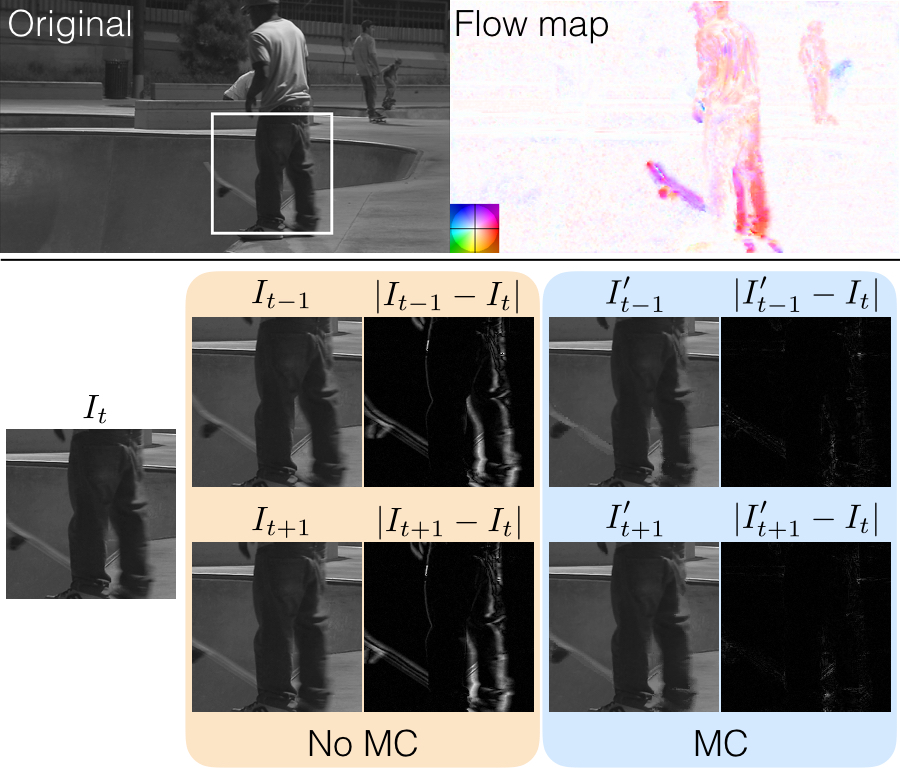}
    \caption{Spatial transformer motion compensation. Top: flow map estimated relating the original frame with its consecutive frame. Bottom: sections of three consecutive frames without and with motion compensation (No MC and MC). Error maps are less pronounced for MC.}
    \label{fig:memc}
\end{figure}

\begin{figure}
  \centering
  \begin{subfigure}[b]{0.32\columnwidth}
      \centering
      \includegraphics[width=\textwidth]{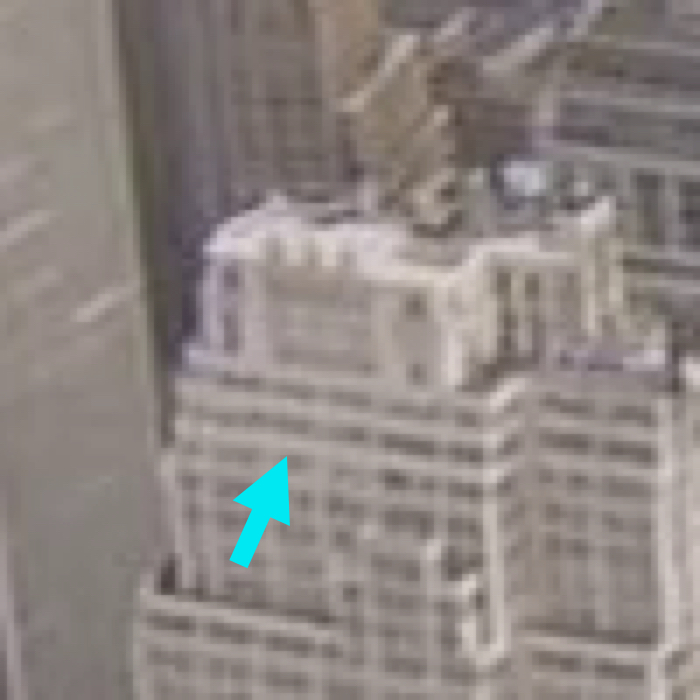}
      \caption{Original}
      \label{fig:mcsr-original}
  \end{subfigure}
  \begin{subfigure}[b]{0.32\columnwidth}
      \centering
      \includegraphics[width=\textwidth]{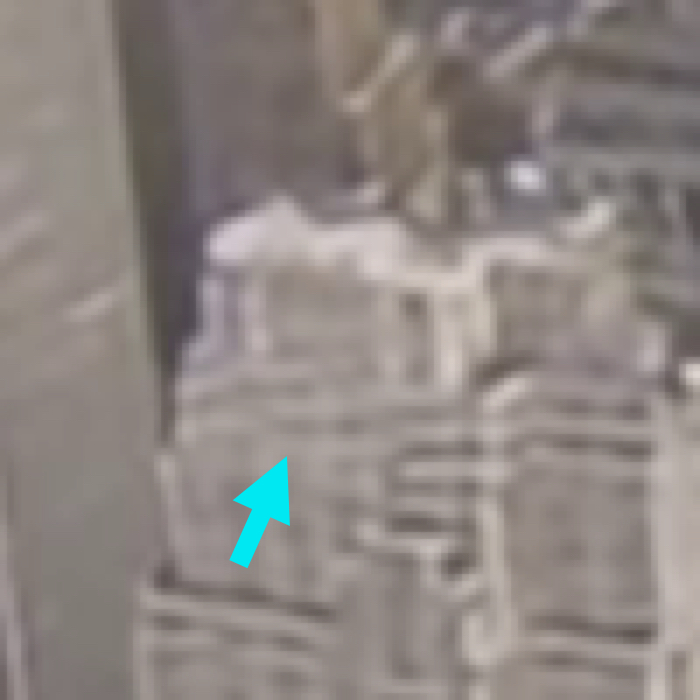}
      \caption{No MC $\times 3$}
      \label{fig:mcsr-no-mc}
  \end{subfigure}
  \begin{subfigure}[b]{0.32\columnwidth}
      \centering
      \includegraphics[width=\textwidth]{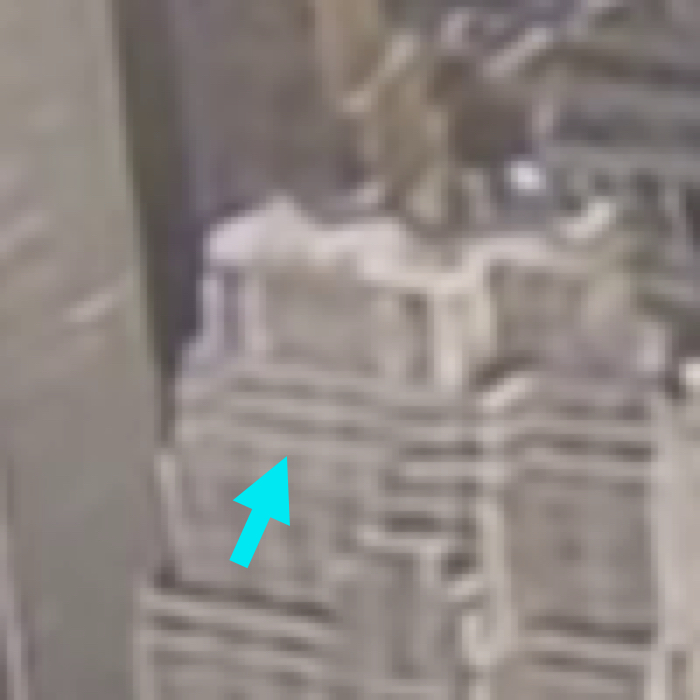}
      \caption{MC $\times 3$}
      \label{fig:mcsr-no}
  \end{subfigure}
  \caption{Motion compensated $\times 3$ \gls{SR}. Jointly motion compensation and \gls{SR} (c) produces structurally more accurate reconstructions than spatio-temporal \gls{SR} alone (b).}
  \label{fig:mcsr-x3}
\end{figure}

In this section, the proposed frame motion compensation is combined with an early fusion network of temporal depth $D_0=3$. First, the motion compensation module is trained independently using \cref{eq:video-sr-memc}, where the first term is ignored and $ \beta = 1$, $\lambda = 0.01$. This results in a network that will compensate the motion of three consecutive frames by estimating the flow maps of outer frames relative to the middle frame. An example of a flow map obtained for one frame is shown in \cref{fig:memc}, where we also show the effect the motion compensation module has on three consecutive frames.



\begin{table}[t]
\centering
\footnotesize
\begin{tabular}{c|cccc}
\# Layers & 6      & 7      & 8       & 9       \\ \hline
SF     & 37.718 & 37.780 & 37.812  & 37.800  \\
E3     & 37.842 & 37.889 & 37.956  & 37.980  \\
E3-MC   & 37.928 & 37.961 & 38.019 & 38.060
\end{tabular}
\caption{PSNR for CDVL $\times 3$ \gls{SR} using single frame (SF) and 3 frame early fusion without and with motion compensation (E3, E3-MC).}
\label{tab:motion-compensated-video-sr}
\end{table}

\begin{table*}[]
\centering
\footnotesize
\begin{tabular}{|c|c|c|c|c|c|c|c|l}
\cline{1-8}
\multicolumn{1}{|l|}{} &                             & \multicolumn{4}{c|}{Image and video SR}               & \multicolumn{2}{c|}{Proposed VESPCN}  &  \\
Scale                  &                                                     & Bicubic   & SRCNN  & ESPCN  & VSRnet  & 5L-E3 & 9L-E3-MC &  \\ \cline{1-8}
3                      & PSNR                                              & 25.38   & 26.56  & 26.97  & 26.64   & 27.05 & \textbf{27.25} &  \\
                       & SSIM                                                & 0.7613  & 0.8187 & 0.8364 & 0.8238  & 0.8388 & \textbf{0.8447} &  \\
                       & MOVIE ($\times 10^{-3}$)               & 5.36       & 3.58 & 3.22 & 3.50           & 3.12        & \textbf{2.86} & \\
                       & GOps / $1080$p frame           						& -       & 233.11 & 9.92   & 1108.73* & \textbf{7.96} & 24.23 &  \\ \cline{1-8}
4                      & PSNR                                              & 23.82   & 24.68  & 25.06 & 24.43   & 25.12 & \textbf{25.35} &  \\
                       & SSIM                                                & 0.6548 & 0.7158 & 0.7394 & 0.7372 & 0.7422 & \textbf{0.7557} &  \\
                       & MOVIE ($\times 10^{-3}$)               & 9.31     & 6.90 & 6.54     & 6.82            & 6.18 & \textbf{5.82}        &  \\
                       & GOps / $1080$p frame 								& -       & 233.11 & 6.08   & 1108.73* & \textbf{4.85} & 14.00 &  \\ \cline{1-8}
\end{tabular}
\caption{Performance on Vid4 videos. *VSRnet does not include operations needed for motion compensation.}
\label{tab:set4}
\end{table*}

\begin{figure*}
\centering
	\includegraphics[width=0.162\textwidth]{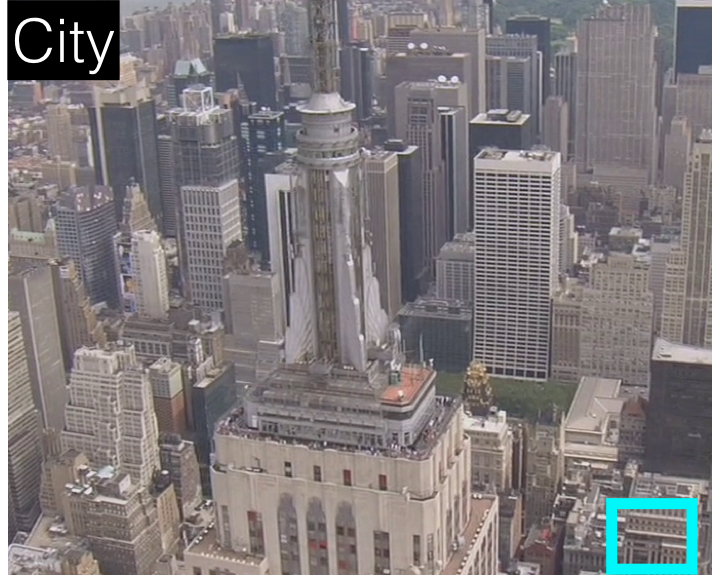}	
	\includegraphics[width=0.162\textwidth]{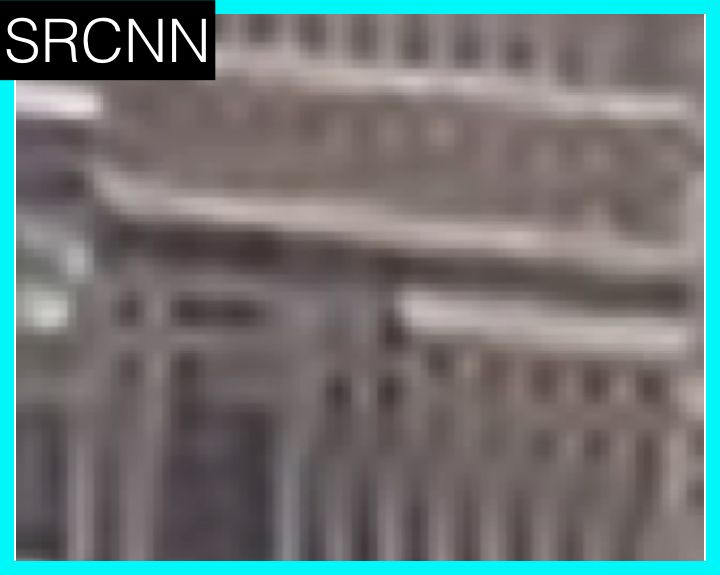}
	\includegraphics[width=0.162\textwidth]{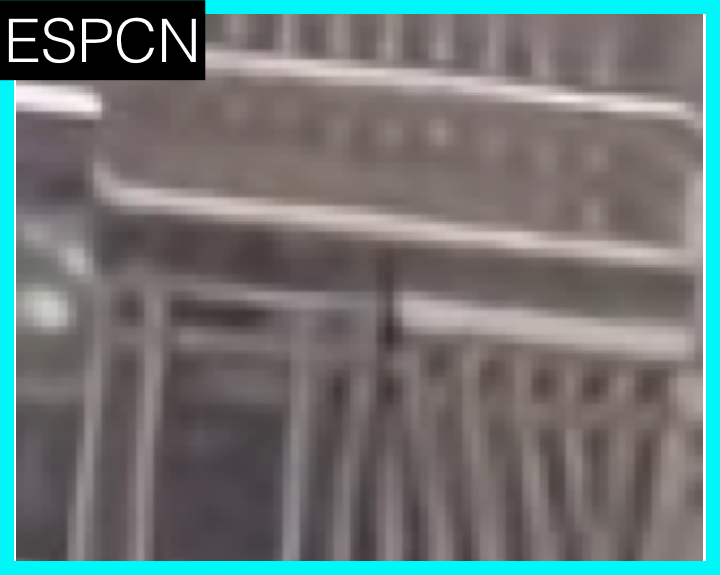}
	\includegraphics[width=0.162\textwidth]{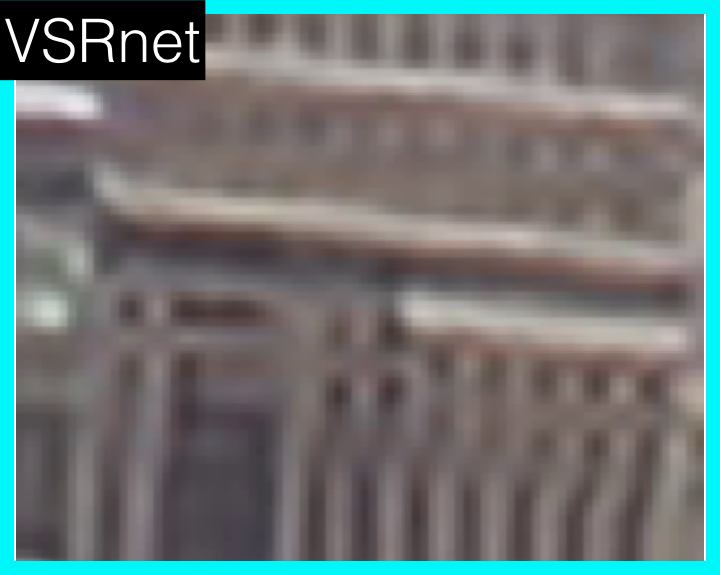}
	\includegraphics[width=0.162\textwidth]{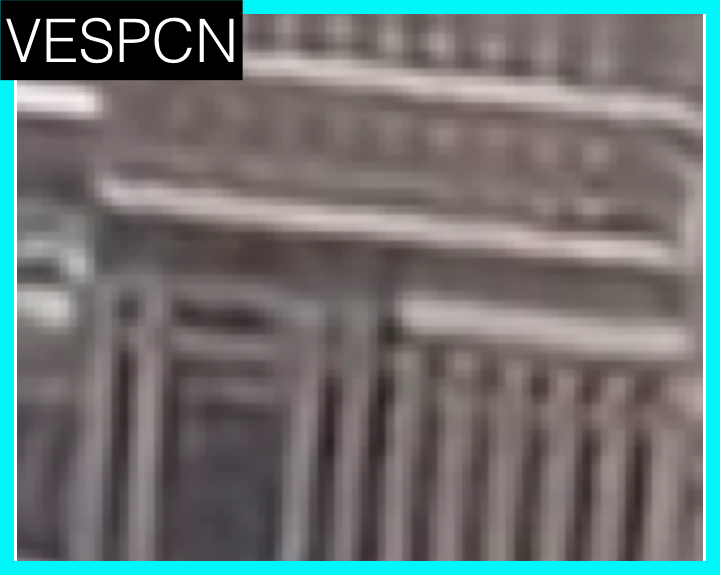}
	\includegraphics[width=0.162\textwidth]{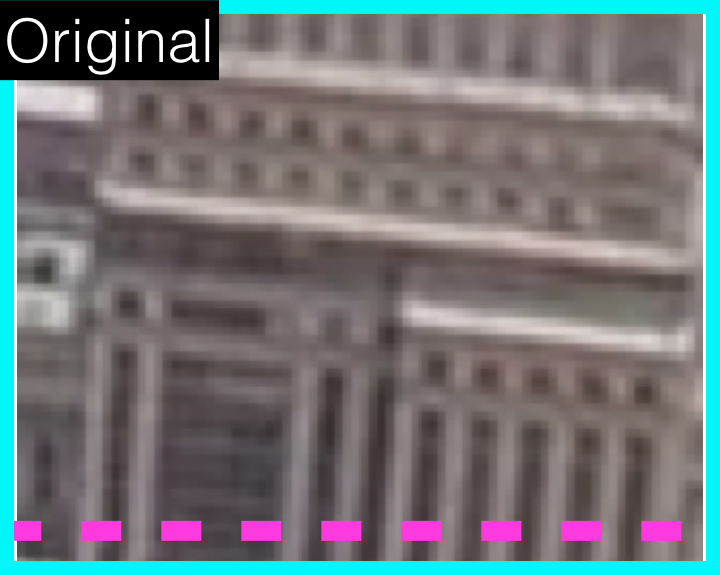}
	
	\includegraphics[width=0.162\textwidth]{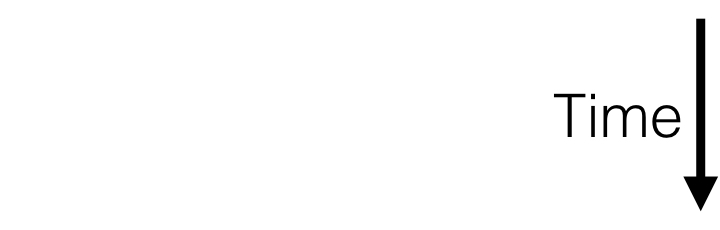}
	\includegraphics[width=0.162\textwidth]{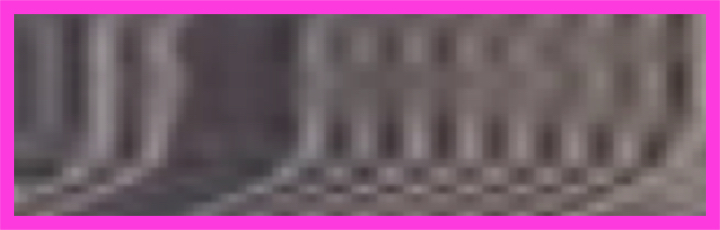}
	\includegraphics[width=0.162\textwidth]{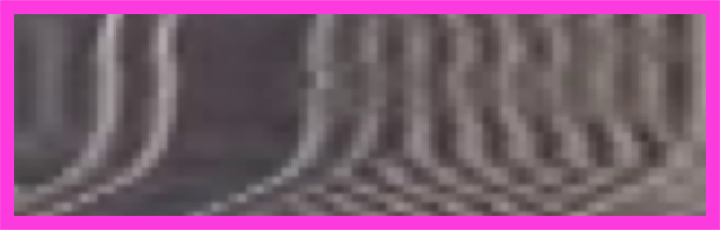}
	\includegraphics[width=0.162\textwidth]{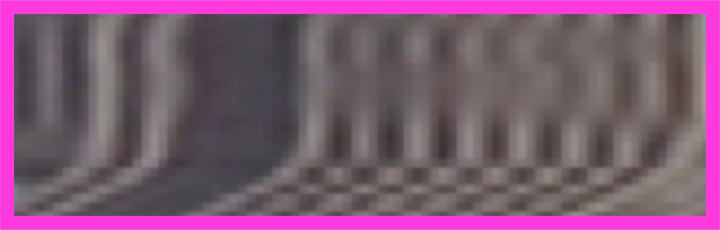}
	\includegraphics[width=0.162\textwidth]{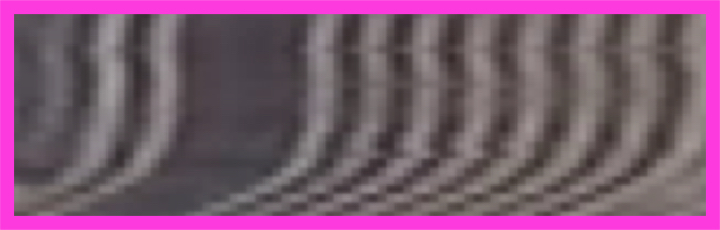}
	\includegraphics[width=0.162\textwidth]{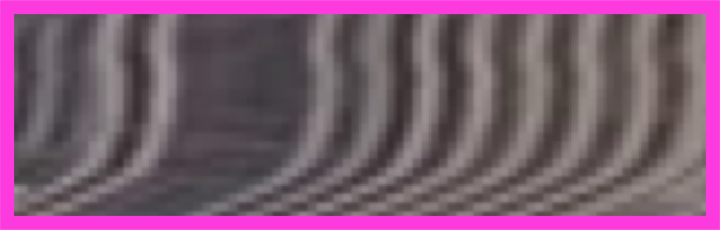}
	
	\vspace{2mm}
	
	\includegraphics[width=0.162\textwidth]{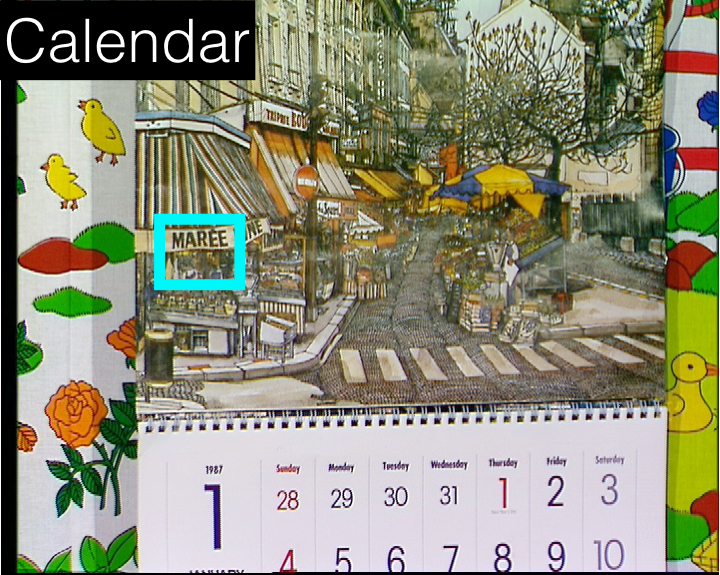}	
	\includegraphics[width=0.162\textwidth]{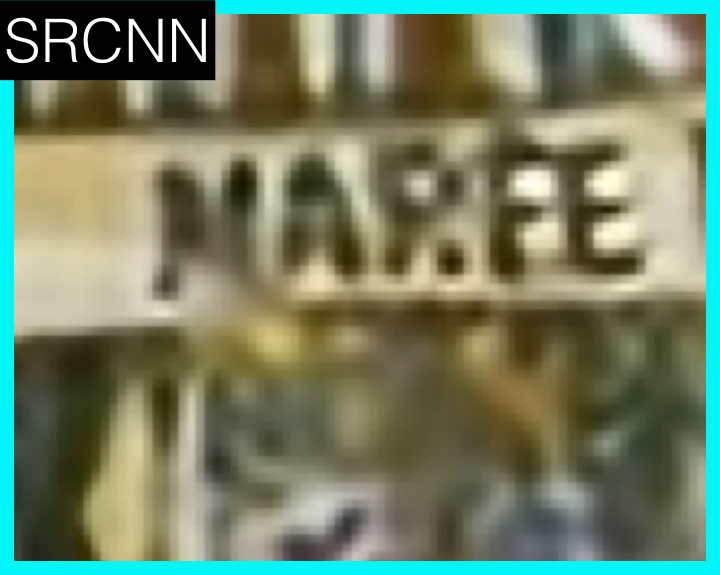}
	\includegraphics[width=0.162\textwidth]{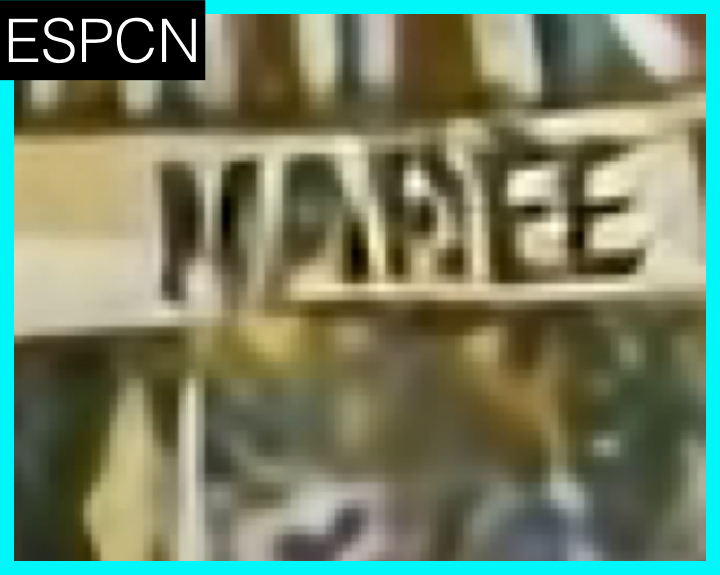}
	\includegraphics[width=0.162\textwidth]{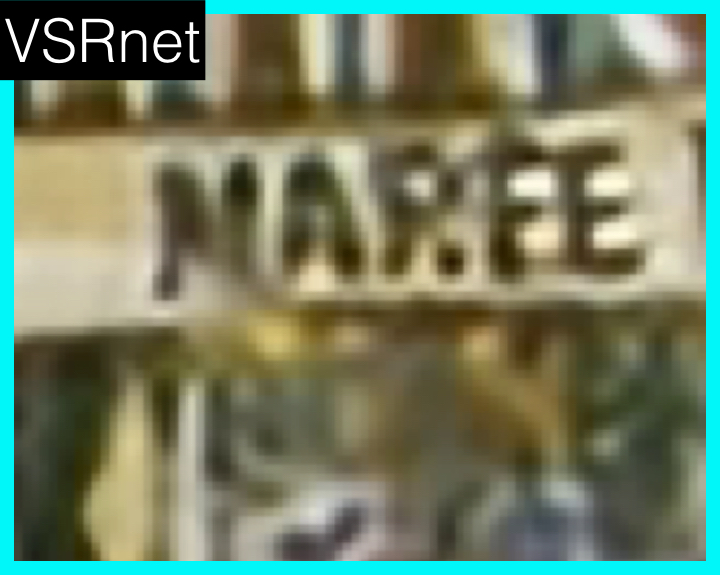}
	\includegraphics[width=0.162\textwidth]{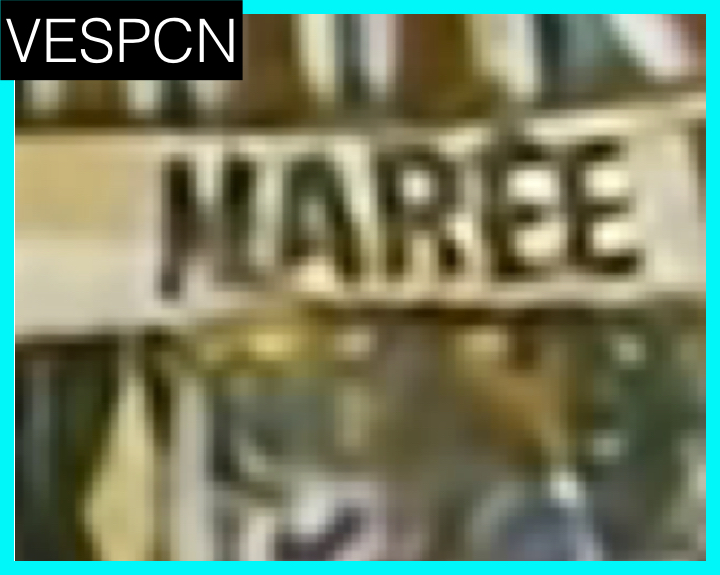}
	\includegraphics[width=0.162\textwidth]{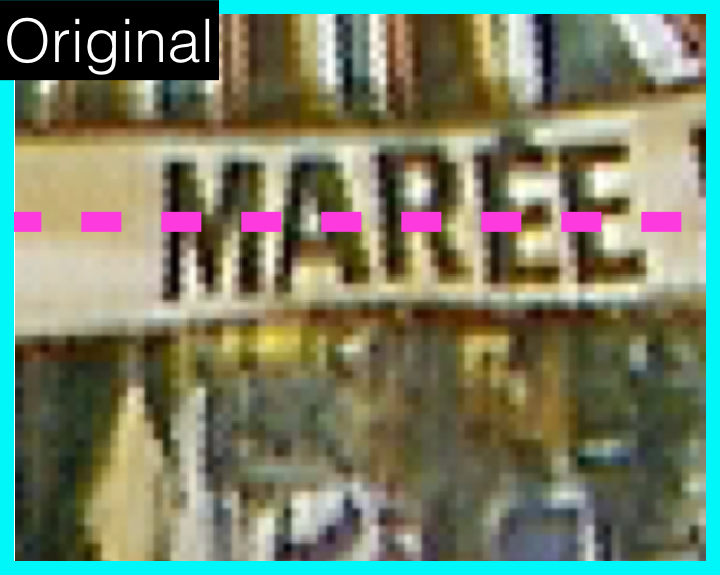}
	
	\includegraphics[width=0.162\textwidth]{crop-profile-axis}
	\includegraphics[width=0.162\textwidth]{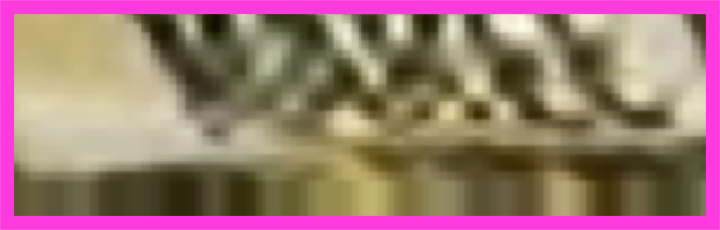}
	\includegraphics[width=0.162\textwidth]{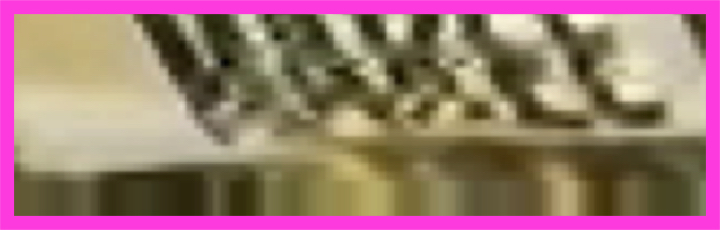}
	\includegraphics[width=0.162\textwidth]{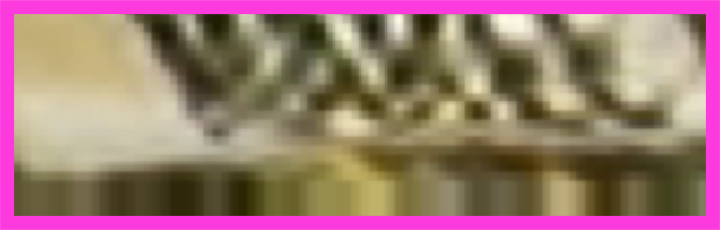}
	\includegraphics[width=0.162\textwidth]{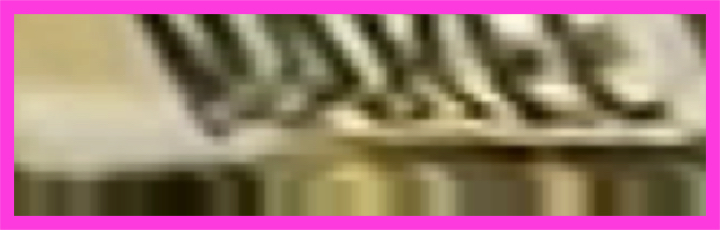}
	\includegraphics[width=0.162\textwidth]{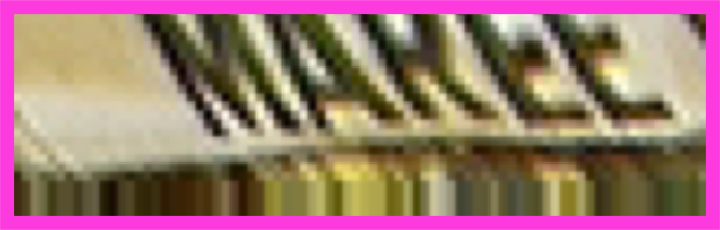}
	
	\caption{Results for $\times 3$ \gls{SR} on Vid4. Light blue figures show results for SRCNN, ESPCN, VSRnet, VESPCN (9L-E3-MC), and the original image. Purple images show corresponding temporal profiles over 25 frames from the dashed line shown in the original image. VESPCN produces visually the most accurate results, both spatially and through time.}
	\label{fig:set4-visualisation}
\end{figure*}

The early fusion motion compensated \gls{SR} network (E3-MC) is initialised with a compensation and a \gls{SR} network pretrained separately, and the full model is then jointly optimised with \cref{eq:video-sr-memc} ($\beta  = 0.01$, $\lambda = 0.001$). Results for $\times 3$ \gls{SR} on CDVL are compared in \cref{tab:motion-compensated-video-sr} against a single frame (SF) model and early fusion without motion compensation (E3). E3-MC results in a \gls{PSNR} that is sometimes almost twice the improvement of E3 relative to SF, which we attribute to the fact that the network adapts the \gls{SR} input to maximise temporal redundancy. In \cref{fig:mcsr-x3} we show how this improvement is reflected in better structure preservation.

\subsection{Comparison to state-of-the-art}

We show in \cref{tab:set4} the performance on Vid4 for SRCNN \cite{Dong2015}, ESPCN \cite{Shi2016}, VSRnet \cite{Kappeler2016} and the proposed method, which we refer to as video ESPCN (VESPCN). To demonstrate its benefits in efficiency and quality we evaluate two early fusion models: a 5 layer 3 frame network (5L-E3) and a 9 layer 3 frame network with motion compensation (9L-E3-MC). The metrics compared are \gls{PSNR}, \gls{SSIM} \cite{Wang2004} and MOVIE \cite{Seshadrinathan2010} indices. The MOVIE index was designed as a metric measuring video quality that correlates with human perception and incorporates a notion of temporal consistency. We also directly compare the number of operations per frame of all \gls{CNN}-based approaches for upscaling a generic $1080$p frame.

Reconstructions for SRCNN, ESPCN and VSRnet use models provided by the authors. SRCNN, ESPCN and VESPCN were tested on Theano and Lasagne, and for VSRnet we used available Caffe Matlab code. We crop spatial borders as well as initial and final frames on all reconstructions for fair comparison against VSRnet \footnote{We used our own implementation of SSIM and use video PSNR instead of averaging individual frames PSNR as done in \cite{Kappeler2016}, thus values may slightly deviate from those reported in original papers.}.

\subsubsection{Quality comparison}

An example of visual differences is shown in \cref{fig:set4-visualisation} against the motion compensated network. From the close-up images, we see how the structural detail of the original video is better recovered by the proposed VESPCN method. This is reflected in \cref{tab:set4}, where it surpasses any other method in \gls{PSNR} and \gls{SSIM} by a large margin. \Cref{fig:set4-visualisation} also shows temporal profiles on the row highlighted by a dashed line through 25 consecutive frames, demonstrating a better temporal coherence of the reconstruction proposed. The great temporal coherence of VESPCN also explains the significant reduction in the MOVIE index.

\subsubsection{Efficiency comparison}


The complexity of methods in \cref{tab:set4} is determined by network and input image sizes. SRCNN and VSRnet upsample \gls{LR} images before attempting to super-resolve them, which considerably increases the required number of operations. VSRnet is particularly expensive because it processes $5$ input frames in $64$ and $320$ feature layers, whereas sub-pixel convolution greatly reduces the number of operations required in ESPCN and VESPCN. As a reference, ESPCN $\times 4$ runs at $29$ms per frame on a K2 GPU \cite{Shi2016}. The enhanced capabilities of spatio-temporal networks allow to reduce the network operations of VESPCN relative to ESPCN while still matching its accuracy. As an example we show VESPCN with 5L-E3, which reduces the number of operations by about 20\% relative to ESPCN while maintaining a similar performance in all evaluated quality metrics.


The operations for motion compensation in VESPCN with 9L-E3-MC, included in \cref{tab:set4} results, amount to $3.6$ and $2.0$ GOps for $\times 3$ and $\times 4$ upscaling, applied twice for each input frame requiring motion compensation. This makes the proposed motion compensated video \gls{SR} very efficient relative to other approaches. For example, motion compensation in VSRnet is said to require 55 seconds per frame and is the computational bottleneck \cite{Kappeler2016}. This is not accounted for in \cref{tab:set4} but is $\times 10^{3}$ slower than VESPCN with 9L-E3-MC, which can run in the order of $10^{-2}$ seconds. The optical flow method in VSRnet was originally shown to run at $29$ms on GPU for each frame of dimensions $512 \times 383$, but this is still considerably slower than the proposed solution considering motion compensation is required for more than a single frame of HD dimensions.

\section{Conclusion}

In this paper we combine the efficiency advantages of sub-pixel convolutions with temporal fusion strategies to present real-time spatio-temporal models for video \gls{SR}. The spatio-temporal models used are shown to facilitate an improvement in reconstruction accuracy and temporal consistency or reduce computational complexity relative to independent single frame processing. The models investigated are extended with a motion compensation mechanism based on spatial transformer networks that is efficient and jointly trainable for video \gls{SR}. Results obtained with approaches that incorporate explicit motion compensation are demonstrated to be superior in terms of PSNR and temporal consistency compared to spatio-temporal models alone, and outperform the current state of the art in video \gls{SR}.

{\small
\bibliographystyle{ieee}
\bibliography{/Users/jcaballero/Documents/Mendeley-bib/CVPR2017}
}

\end{document}